%% file: root.tex
\documentclass[letterpaper, 10 pt, conference]{ieeeconf}  

\IEEEoverridecommandlockouts                              

\usepackage{graphics}
\usepackage{amsmath}
\usepackage{amssymb}
\usepackage{bbm}
\usepackage{mathtools}
\usepackage{algorithm}
\usepackage{algpseudocode}
\usepackage[]{mdframed}
\usepackage{hyperref}
\usepackage{xcolor}
\usepackage{nicefrac}
\usepackage{booktabs}
\usepackage{siunitx}
\usepackage{diagbox}
\usepackage{graphicx}
\usepackage{subcaption}
\usepackage{cite} 

\newcommand{\R}{\mathbb{R}}
\newcommand{\E}{\mathbb{E}}
\newcommand{\cb}{\mathcal{B}}
\newcommand{\cg}{\mathcal{G}}
\newcommand{\cu}{\mathcal{U}}

\DeclarePairedDelimiter\abs{\lvert}{\rvert}

\title{\LARGE \bf Fast Confidence-Aware Human Prediction via Hardware-accelerated Bayesian Inference for Safe Robot Navigation}

\author{Michael Lu$^{1}$, Minh Bui$^{1}$, Xubo Lyu$^{1}$, and Mo Chen$^{1}$
\thanks{$^{1}$School of Computing Science, Simon Fraser University, Canada {\tt\small \{michael\_lu\_3, buiminhb, xly, mochen\}@sfu.ca}}%
}

\begin{document}

\maketitle

\input{0_abstract}
\input{1_introduction}
\input{2_problem_formulation}
\input{3_methodology}
\input{4_experiments}
\input{5_conclusion}

\bibliographystyle{IEEEtran}
\bibliography{ref}

\end{document}

%% file: 0_abstract.tex
\begin{abstract}
As robots increasingly integrate into everyday environments, ensuring their safe navigation around humans becomes imperative.
Efficient and safe motion planning requires robots to account for human behavior, particularly in constrained spaces such as grocery stores or care homes, where interactions with multiple individuals are common. 
Prior research has employed Bayesian frameworks to model human ``rationality'' based on navigational intent, enabling the prediction of probabilistic trajectories for planning purposes.
In this work, we present a simple yet novel approach for confidence-aware prediction
that treats future predictions as particles
This framework is highly parallelized and accelerated on an graphics processing unit (GPU).
As a result, this enables longer-term predictions at a frequency of 125 Hz and can be easily extended for multi-human predictions.
Compared to existing methods, our implementation supports finer prediction time steps, yielding more granular trajectory forecasts. 
This enhanced resolution allows motion planners to respond effectively to subtle changes in human behavior.
We validate our approach through real-world experiments, demonstrating a robot safely navigating among multiple humans with diverse navigational goals. Our results highlight the method’s potential for robust and efficient human-robot coexistence in dynamic environments.
\end{abstract}

%% file: 1_introduction.tex
\section{INTRODUCTION}
A key factor for the adoption of mobile robots in common applications such as delivery or cleaning robots, is the ability to safely and efficiently navigate around humans. 
One important factor that would enable such applications is accurate, robust, and real-time predictions of human motions and trajectories. 
It is essential for human-aware robotic navigation methods to be fast enough in to run in real time~\cite{kruse2013human,moller2021survey} to adapt to dynamic human movement, especially when the robot itself can move at considerable speeds. 
While there has been many works addressing the accuracy of predicting future trajectories, robustness and speed have not been as adequately addressed.

The considerable body of work in the human navigation prediction literature that push further prediction accuracy can be classified as follows. \\
\textbf{Rule based methods.}

This body of research assumes people move according to some predefined models. 
One of earliest methods and most popular method is the Social Force Model \cite{helbing1995social}. 
Under this regime, each human is modeled as a particle with a certain mass.
Attractive force and repulsive force come from the goal reaching intention and interacting with other agents and obstacles.
Due to its simplicity, the method has been influential to many later works that built upon this idea \cite{SFM2, SFM3, SFM4}. 
Other compelling methods assumes that an humans in a group dynamic will always choose to cooperate to avoid collision with each other \cite{BRVO, freezingrobot}.
Game-theoretic principles have also been applied to solve for an Nash equilibrium to generate socially compliant trajectories for all agents including the robots \cite{iLQR, Sun2024MixedStrategyNE}. 
The deterministic nature of these approaches means that they are usually easy to implement and can be deployed real-time applications. 
However, humans may not always follow these rules or principles.

\textbf{Data-driven eased methods.}
Instead of hardcoding the rules for human movement, human trajectories can be learned from data.
For example, inverse reinforcement learning has been used to learn human preferences when navigating in an environment \cite{IRLplanning}.
More recently, the success of data-driven methods has inspired the use of many deep learning architectures such as long short term memory (LSTM), generative adversarial networks (GANs), transformers, diffusion models to learn to predict human trajectories from large public datasets~\cite{socialGan,Alahi2016, zhitianDiffusion, Salzmann2020TrajectronDT}.
However, these methods often require significant computational power for training and deployment.
Additionally, predicting trajectories in real-time may not be feasible due to long inference time of large networks.
More importantly, the black-box nature of these approaches make it hard to deploy these methods in safety-critical situations.

\textbf{Confidence-based Human Predictions}
A large majority of the work discussed previously does not take into account uncertainty of their predictions, which is crucial in safety-critical situations.
It is equally important, if not more so, that the robot's prediction is aware of an human's changing interest in order to plan safely. 
for example, a person can suddenly stop to check their phone, thus changing the probability distributions over their future states.
a recent line of research employed confidence-aware predictions~\cite{fisac2018probabilistically,fridovich2020confidence} using a bayesian framework to model the future trajectories of a human.
This framework is highly adaptable and robust, allowing for the adjustments to subsequent prediction when a human may unexpectedly deviate from their predicted trajectory.
Additionally this approach is simple since no prior trajectories are required and only a mild number of assumptions (such as possible goal locations) are needed.
Moreover, this approach has been successfully used for real-time multi-robot, multi-human navigation~\cite{bajcsy2019scalable} and have been further extended to also incorporate a human's changing navigational intent and their urgency in quickly reaching their goal~\cite{agand2022human}.

A caveat to these approaches is that to enable real world demonstrations, the robot moved slowly due to slow real-time predictions.
Additionally, only a small (at most 2) number of humans was considered~\cite{bajcsy2019scalable}.
These limitations are a result of the use of full Bayesian inference, which necessitates iterating over all possible human states.
A partial sampling-based approach was proposed in~\cite{agand2022human}, but the approach could not be run in real time.

\textbf{Contributions}: 
We fully parallelize the prior Bayesian framework for human-prediction by representing all distributions using particles for real-time prediction on an GPU.
Our approach enables longer-term predictions with finer prediction time steps, which results in a 300 times speed up compared to equivalent approaches on an CPU.
As a result of faster predictions, our approach also easily supports the prediction for multiple humans.
Using fast real-time predictions, we demonstrate safe, agile robot planning by integrating the predictions to an anytime time-varying A$^*$ planner and a Model Predictive Path Integral (MPPI) controller in a constrained environment with multiple people.

%% file: 2_problem_formulation.tex
\section{PROBLEM FORMULATION}

Consider an robot with state $z^R \in \R^{n_R}$ and control $u^R \in \R^{m_R}$ that evolves according to the system dynamics $\dot{z}^R = f^R(z^R, u^R)$. 
Given a sequence of human states collected at discrete intervals, $z_1, z_2, \dots, z_t$, the objective to predict the human's next $T$ states, $z_{t+1}, z_{t+2}, \dots z_{t + T}$.
For convenience, let $z_{t_1:t_2}$ denote the sequence of states from time $t_1$ to $t_2$ with an interval of $\Delta t$.
Since the number of feasible states can be large, we will instead model the next predicted state $z_{t+1}$ given a sequence of observed states $z_{1:t}$, i.e, $p(z_{t+1} \, | \, z_{1:t})$.

To ensure this distribution can be computed efficiently we will use a non-parametric representation of model the distribution, e.g, an occupancies grid.
We emphasize the simplicity in this approach since the predictions only requires knowledge of a human's previous states and does not depend on any external dataset.
Given that a human may suddenly change their behaviours, it is necessary to quickly adapt to their new preferences with new predictions. 
To do so, we will generate confidence-aware human navigational predictions that can be used for robot planning and will explain how to in the following section.

%% file: 3_methodology.tex
\section{METHODOLOGY}
To achieve real-time, robust prediction of human future state probability distributions without assuming access to any training data, we will build on works such as \cite{fisac2018probabilistically,fridovich2020confidence,baker2007goal,agand2022human} and take on a control theoretic approach.

\subsection{Model-Based Confidence-Aware Human Prediction}
Let $z \in \R^{n_H}$ be the human state, with its time evolution modeled by a dynamical system $f$:

\begin{equation}
	\dot{z} = f(z, u),
\end{equation}

\noindent where the control $u \in \R^{m_H}$ represents the human navigational decision at any given time.
In practice, we only have access to the observed humans states $z_{1:t}$ observed states discrete time $1, 2, \dots, t$. The dynamics $f$ must be \textit{differential flat}~\cite{fliess1995flatness}.\footnote{A dynamical system is differentially flat if there exists flat outputs of the form $y = \alpha(z, u, \dot{u}, \dots, u^p)$ such that $z = \beta(y, \dot{y}, \dots, y^q)$ and $u = \gamma(y, \dot{y}, \dots, y^r)$.} so that the control can be recovered from solely from state observations.\footnote{We note that practical requirement has not been explicitly stated in prior work.}

\begin{mdframed}
\textit{Running Example}: We will use choose $f$ to be the following 2D-kinematic model:
\begin{equation}\label{eq:human_dynamics}
	\dot{z} = \begin{bmatrix}\dot{x} \\ \dot{y} \end{bmatrix}= f(z, u) = \begin{bmatrix}
		v \cos(\theta) \\ v \, \sin(\theta)
	\end{bmatrix},
\end{equation}
\noindent where $z = [x, y]^\top$ is the position of the human, $u = [v, \theta]^\top$ the control consisting of the movement speed $v$ and movement direction $\theta$.
\end{mdframed}

Furthermore, let $\cu \subseteq \R^{m_H}$ be defined as the set of possible human's control under the assumption that they are discrete and bounded. 

Following~\cite{fisac2018probabilistically,fridovich2020confidence,baker2007goal,agand2022human}, we will assume that the human intends to reach some unknown goal location $g \in \R^{n_H}$ using a policy $\pi(\cdot \, | \,  z; \beta, g)$ given by a Boltzmann distribution

\begin{equation}\label{eq:human_softmax_policy}
	\pi(u \, | \,  z; \beta, g) = \frac{\exp(\beta \, Q_H(z, u; g))}{\sum_{u'} \exp(\beta \, Q_H(z, u'; g))},
\end{equation}

\noindent where $Q_H: \R^{n_H} \times \R^{m_H} \to \R$ is the \emph{unknown state-control value} and $\beta > 0$ is the \textit{rationality coefficient}.
This policy represents a noisy-rational model of a human's behaviour where a human is more likely to make decisions that result in higher utility $Q_H$.
The function $Q_H$ models what utility a human expects to receive when taking control $u$ at a particular state $z$ conditioned on the goal $g$.
For example the utility function can be estimated using collected data of human behaviour or based on reachability, such as a time-to-reach value function~\cite{agand2022human}.
The $\beta$ term models how human is expected to behave:
A larger $\beta$ models more rational human behavior that solely aims to maximize the utility function; a smaller $\beta$ models the human's behaviour to be uncertain, and converges towards as uniformly random policy as $\beta \to 0$.
An alternative interpretation of $\beta$ is that it represents the confidence of the human behavioral model $\pi(\cdot \, | \, z; \beta, g)$.
Finally, let $\cg$ represent a finite set of the human's goals, and let $\cb$ represent finite set of rationality coefficients $\beta$ a human could have.

\begin{mdframed}
\textit{Running Example}: Consider a human shopping in a grocery store.
Let $\cg := \{g_i\}_{i=1}^n \subseteq \R^2$ represent the location of grocery items within the store. 
For each $g_i \in \cg$, let $Q_H(z, u; g_i) := -\|z - g_i\|_2^2  - \|u\|^2$ model the human's desire to reach their goal as quickly, yet efficiently, as possible. 
Let $\cb$ represent an discrete random variable with support $\{0.1, 0.5, 1. 5\}$. 
\end{mdframed}

Over time, a human's preference and rationality may change based on environmental factors or changing interests. 
Hence, following~\cite{fridovich2020confidence}, we will model the \emph{rationality coefficient} and \emph{predicted goals} as an hidden Markov model where the belief of the distribution $\cb \times \cg$ will be updated using online data.
For each $(\beta, g) \in \cb \times \cg$, given a history of previous human states $z_{1:t}$ and the new observed state $z_{t+1}$, the Bayesian update of the joint distribution $\beta, g$ is

\begin{equation}\label{eq:beta_update}
	b^{t+1}(\beta, g) = \frac{\pi(u_t \, | \, z_t; \beta, g) \, b^t(\beta, g)}{\sum_{(\beta', g') \in \cb \times \cg} \pi(u_t \, | \,  z_t; \beta', g') \, b^t(\beta', g')} \, ,
\end{equation}

\noindent where $b^t(\beta, g) := p(\beta , g \, | \,  z_{1:t})$, $\pi(\cdot \, | \,  z_t; \beta, g)$ is defined 
in~Eq.\eqref{eq:human_softmax_policy} and $u_t$ is the recovered from $z_t$ and $z_{t+1}$ via the differential flatness of $f$. 
For convenience, let $b^t(\cb, \cg) := \{b^t(\beta, g) \, | \, (\beta, g) \in \cb \times \cg\}$ denote the set of beliefs over all $(\beta, g)$ pairs.

All together, given the history of the human's previous states $z_{1:t}$, the distribution of the human's future occupancies can be computed as follows.

\begin{subequations}\label{eq:human_prediction}
\begin{align}
	&p(z_{t+1} | z_{1:t}) \notag \\ &= \sum_{u, \beta, g} p(z_{t+1}, u, \beta, g \, | \,  z_{1:t}) \\
	&= \sum_{\beta, g} p(\beta, g \, | \,  z_{1:t}) \sum_{u} p(z_{t+1}, u \, | \,  z_{1:t}, \beta, g)  \\
	&= \sum_{\beta, g} b^{t+1}(\beta, g) \sum_{u} p(z_{t+1} \, | \,  z_t, u; \beta, g) \, \pi(u \, | \,  z_t; \beta, g) \label{eq:human_prediction_beta_update}.
\end{align}
\end{subequations}

If the human's dynamics are deterministic and Eq.~\eqref{eq:human_dynamics} discretized, $p(z_{t+1} \, | \,  z_{1:t}; u, \beta, g) = \mathbbm{1}\{z_{t+1} = z_t + f(z_t, u) \Delta t\}$ where $\mathbbm 1$ is the indicator function and $\Delta t$ is the duration of each discretized time step. 
For longer predictive horizons, the future states can estimated by applying Eq.~\ref{eq:human_prediction} iteratively,
\begin{subequations}\label{eq:human_prediction_multi_step}
\begin{align}
p(z_{t+2} | z_t) &= \sum_{z_{t+1}}p(z_{t+2}, z_{t+1} | z_t) \\  &=\sum_{z_{t+1}}p(z_{t+2} | z_{t+1}) p(z_{t+1} | z_t)	
\end{align}
\end{subequations}

In the following section, we will discuss how Eqs.~\eqref{eq:human_prediction}and ~\eqref{eq:human_prediction_multi_step} can be approximated via sampling to facilitate GPU parallelization, enabling fast online prediction for real-time systems.

\subsection{Parallelizable Human-Prediction}
In order to predict a human's future $T$ occupancies, computing Eq.~\eqref{eq:human_prediction}  
requires $\mathcal{O}(T \, \abs{\cb} \, \abs{\cg} \, \abs{\cu})$ operations since we must iterate over each $(u, \beta, g)$ for each subsequent time step. 
Additionally, the space required to store the prediction is $\mathcal{O}(\abs{X} \, \abs{Y})$ where $X$ and $Y$ represent the set of finite grid points the human could be located at. 
When either $T$, $\abs{X}$, $\abs{Y}$ is large, fully computing Eq.~\eqref{eq:human_prediction} in real time may be infeasible for even a single human.

To mitigate this computational challenge, we adopt a particle-based representation to approximate the probability distributions of predicted states. 
This approach is widely used in robotics for planning and prediction tasks in uncertain environments, as demonstrated in prior work~\cite{thrun2002particle,montemerlo2002conditional}.
Observe that Eq.\eqref{eq:human_prediction_beta_update} can equivalently as

\begin{subequations} \label{eq:expectation}
	\begin{align}
		&p(z_{t+1} | z_{1:t}) \notag \\ 
		&= \sum_{\beta, g} b^{t+1}(\beta, g) \sum_{u} p(z_{t+1} \, | \,  z_t, u; \beta, g) \, \pi(u \, | \,  z_t; \beta, g) \\
		&= \E_{(\beta, g) \sim b^{t+1}(\beta, g)} \left[\sum_{u} p(z_{t+1} \, | \,  z_t, u; \beta, g) \, \pi(u \, | \,  z_t; \beta, g) \right]\tag {Definition of an expectation of a discrete r.v.}\\
		&= \E_{(\beta, g) \sim b^{t+1}(\beta, g)} \left[\E_{u \sim \pi(u \, | \,  z_t; \beta, g)}\left[p(z_{t+1} \, | \,  z_t, u; \beta, g) \,  \right]\right] \tag {Definition of an expectation of a discrete r.v.}
		\\
		&= \E_{(\beta, g) \sim b^{t+1}(\beta, g)} \left[ \right . \notag \\
		& \left.  \quad \ \ \E_{u \sim \pi(u |, z_t; \beta, g)}\left[\E_{z_{t+1} \sim f(z_t, u)}[\mathbbm{1}\{z_{t+1} = f(z_t, u)\}]\right]\right] \notag \\
		&= \E_{\substack{(\beta, g) \sim b^{t+1}(\beta, g) \\ u \sim \pi(u |, z_t; \beta, g) \\ z_{t+1} \sim f(z_t, u)}}[\mathbbm{1}\{z_{t+1} = f(z_t, u)\}]\label{eq:human_prediction_expectation}. 
	\end{align}	
\end{subequations}

\noindent where second last equality obtained since probability of an event is equal to the expectation of the indicator of the event, and the last equality is due to the law of total expectation, where the joint expectation is over all possible $(\beta, g, u, z_{t+1})$ tuples.
As a result, Eq. \eqref{eq:human_prediction_expectation} can be approximated by an Monte-Carlo approximation by using a sufficiently large number of samples.
To do so, we will represent states as particles and evolving them using the weights from the above equation. For example, given an initial particle at state $\tilde{z}_t$, we can sample an $(\beta, g)$ pair from $b^{t}(\cb, \cg)$, sample control $\tilde{u} \sim \pi(\cdot \, | \, z_t; \beta, g)$ and obtain new state $\tilde{z}_{t+1} = \tilde{z}_t + f(\tilde{z}_t, \tilde{u})\Delta t$.
A large number of particles are needed to sufficiently represent the probability distribution.
To mitigate this challenge, will evolve them in parallel on an GPU.
Alg. \ref{alg:hardware_accelerated_human_prediction_code} provides the pseudo-code for approximating Eq.~\eqref{eq:human_prediction}.

\begin{algorithm}
\caption{Hardware-Accelerated Human Prediction}\label{alg:hardware_accelerated_human_prediction_code}
\begin{algorithmic}[1]
\Require Number of samples $n$, predictive horizon $T$, prediction time step $\Delta t$, initial states $z_{1:t'}$, beliefs $b^{t'}(\cb, \cg)$
\State Create $n$ duplicate particles of $z_{t'}$ and create batch $[\tilde{z}_{1}]_{i=1}^n$
\State Sample $n$ behavioural and goals pairs $([\beta]_{i=1}^n, [g]_{i=1}^n) \sim b^{t'}(\cb, \cg)$ 
\For{prediction steps $t=1$ to $T$}
	\State \textit{// Executed in parallel for each particle $i$}
	\For{$(\tilde{z}_t, \beta, g) \in [\tilde{z}_{t}]_{i=1}^n \times [\beta]_{i=1}^n \times [g]_{i=1}^n$} \label{alg:parallel_loop}
		\State $u_t \sim \pi(\cdot \, | \,  \tilde{z}_t; \beta, g)$ using Eq.~\ref{eq:human_softmax_policy}
		\State $\tilde{z}_{t+1} \gets \tilde{z}_t + f(\tilde{z}_t, u_t) \Delta t$
	\EndFor
	\State Emplace $[\tilde{z}_{t+1}]_{i=1}^n$ onto a grid to obtain $p(z_{t' + t} | z_{1:t'})$
\EndFor
\Ensure $p(z_{t' + t} \, | \,  z_{1:t'})$ for each $t=1 \dots T$
\end{algorithmic}
\end{algorithm}
\noindent For the initial time step, $t=1$, we initialize the joint belief distributions as $b^1(\beta, g) = \frac{1}{\abs{\cb} \, \abs{\cg}}$. 
Unlike in~\cite{agand2022human}, our approach leverages propagating particles in parallel on a GPU, allowing for significantly more particles to be used when generating predictions, leading to more accurate predictions in the form of arbitrary (e.g. multi-modal) distributions over future human states.

According to Eq.~\ref{eq:human_prediction_beta_update}, we must compute $b^{t+1}(\beta, g) = p(\beta, g \, | \, z_{1:t+1})$ for each future time step. 
However, updating $b^{t+1}(\beta, g)$ in parallel is challenging due to the need to maintain consistent distributions across each goal and behavior. 
To address this issue, we employ a ``bootstrapped'' approach, using the initial estimates $b^{t'}(\beta, g)$ to approximate the update beliefs in each prediction step.
This is a common approach taken in prior work~\cite{fisac2018probabilistically,agand2022human} 
and avoids the assumption of conditional independence between behaviors, i.e. $p(\beta, g | z_{1:t}) = p(\beta | z_{1:t}) p(g | z_{1:t})$. 
By leveraging faster predictions, our method captures more complex human navigational behaviors, as each goal $g$ is associated with its own set of behaviours.

To convert the particles into probability mass, we will use a point estimate when mapping the particles onto a discrete grid.
To ensure a smooth distribution, we apply a Gaussian convolution to the batch of particles $[z_{t }]_{i=1}^n$ in each iteration before emplacing them onto a a grid. 
This convolution can be done efficiently on the GPU. 

We observed that this convolution significantly improves the continuity and accuracy of the predictions, since the smoothing step helps eliminate gaps when the particles are mapped onto a grid.
Notably, the particles in each iterations only placed onto a grid when estimating $p(z_{t+1} | z_{t})$
This prevents the unnecessary discretization of particles across multiple predictions.
In practice, it is common to store the resulting predictions on a grid as it is a common input for trajectory planners.

For observations $z_t$ and $z_{t+1}$, the corresponding control $u_t$ can be recovered because the assumed human model $f$ is differentiable flat.
As previously discussed, paralleling the updates in Eq.~\eqref{eq:beta_update} is difficult due to their interdependence.
Nevertheless, we can still perform the updates on the GPU to avoid the computation overhead associated with frequent data transfer between the GPU and CPU.
To prevent numerical instabilities, we store and update the beliefs in the log-probability space in the following equation: 
\begin{subequations}
	\begin{align}
		&\log b^{t+1}(\beta, g) \\ &= \log\left(\frac{\pi(u_t | z_t; \beta, g) \, b^t(\beta, g)}{\sum_{(\beta', g') \in \cb \times \cg} \pi(u_t \, | \,  z_t; \beta', g') \, b^t(\beta', g')}\right) \\
		&= \log(\pi(u_t | z_t; \beta, g)) + \log(b^t(\beta, g)) \notag \\&\quad - \log\left(\sum_{(\beta', g')} \pi(u_t \, | \,  z_t; \beta', g') \, b^t(\beta', g')\right).
	\end{align}
\end{subequations}
The final term, a LogSumExp operation, is computed efficiently and only once during belief updates.

Given an discrete occupancy grids $\{p(z_{t' + t} \, | \,  z_{t'})\}_{t=1}^T$ representing the probability at a human is at state $z$ at time $t$, we will next discuss how these predictions can be integrated in time-varying planner to generate safe robotic trajectories. 

\subsection{Safe Trajectory Planning}\label{sec:safe_traj_planners}
The human predictions can be integrated as an obstacle map or occupancy grid into either time-varying global or local planners for safe planning. In both approaches, we will replan frequency to take advantage of the fast predictions.
\begin{mdframed}
\textit{Running Example}: We will the model the dynamics of the robot $f^R$ to be the following 4D-Dubins car model:
\begin{equation}\label{eq:robot_model}
	\dot{z}^R = \begin{bmatrix}\dot{x} \\ \dot{y} \\ \dot{v} \\ \dot{\theta}\end{bmatrix} = f^R(z^R, u^R) = \begin{bmatrix}
		v \cos(\theta) \\ v \, \sin(\theta) \\ a \\ \omega
	\end{bmatrix}.
\end{equation}

Here, $z^R = [x, y, v, \theta]^\top$ is the robot state, with $(x, y)$ being the position, $v$ the speed, and $\theta$ the heading. 
The control inputs are $u^R = [v, \omega]^\top$ where $a$ is acceleration and $\omega$ is the steering angle.
\end{mdframed}

\subsubsection{Global Planner}\label{sec:ana}
To formulate a global planner we use Anytime Nonparametric A$^*$ (ANA$^*$)~\cite{van2011anytime}, an anytime time-varying A$^*$ algorithm. 
The anytime property is necessary due to changing obstacles in the environment, i.e the human predictions.
ANA$^{*}$ does not require explicitly depend on the time of the goal state unlike more common anytime search planners such as Anytime Repairing A$^*$~\cite{likhachev2003ara} 
or Anytime Dynamic A$^*$~\cite{likhachev2005anytime}.
This trait is needed since the predicted time of when the robot will reach the goal is unknown due to the human predictions.
\begin{mdframed}
\textit{Running Example}: We use the following cost for ANA$^*$ to integrate the predictions into the planning,
\begin{align}
	c_{\text{ANA*}}\left(z^R_t, u^R_t, t\right) &= 1000 \, \* \, \mathbbm{1} \{\text{coll}(z_t^R, z_t) \geq 0.1 \} \notag \\ &\quad + \Delta_\text{plan} t
\end{align}
where $\text{coll}(z_t^R, z_t)$ denotes the probability that robot collides with a predicted human state at time $t$, $\Delta_{\text{plan}} t$ is the planning interval and use $-\|z^R - g^R \|^2$ as the heuristic and $g^R$ is the goal state of the robot.
\end{mdframed}
Once a collision free path is returned or planning time-limit has been reached an linear quadratic regulator (LQR) controller is used to follow the path.
\subsubsection{Local Planner}\label{sec:mppi}
To demonstrate the advantages of fast predictions, we will use an Model Predictive Path Integral (MPPI)~\cite{williams2016aggressive,williams2018information} controller.
As a brief overview, given initial state $z^R_{t'}$ and a sequence of discrete control references $\{u^R_{t' + t}\}_{t=1}^K$, $N$ random control perturbations are sampled
$[\{\delta_{t'=t}\}_{t=1}^K]_{i=1}^N$ and is applied to each control sequence, i.e 
$\{u^{R'}_{t' + t}\}_{t=1}^K = \{u^R_{t' + t} + \delta_{t' +t} \}_{t=1}^T$. Then the trajectory cost $S(\xi^i_{t' + t})$ when staring at initial state $z_{t' + t}$ is computed for each generated control sequence. Finally, the optimal control sequence for each $t=1, 2 \dots, T$ can be approximated as~\cite{williams2017model}
\begin{equation}
	u^*_{t' + t} = u^R_{t' + t} + \frac{\sum_{i=1}^N \exp(-(\nicefrac{1}{\tau}) S(\xi^i_{t' + t})) \, \delta^i_{t' + t}}{\sum_{i=1}^N \exp(-(\nicefrac{1}{\tau}) S(\xi^i_{t' + t}))},
\end{equation}
where $\tau \geq 0$ denotes the ``temperature'' which controls which costs to prioritize.
\begin{mdframed}
\textit{Running Example}: We use the following quadratic cost for MPPI, 
\begin{align}
	c_{\text{MPPI}}\left(z^R_t, u^R_t, t\right) &= (z^R_t - g^R)^\top Q \, (z^R_t - g^R) + R^\top u_t \notag \\ &\quad +c_{\text{ANA*}}(z^R_t, u_t^R, t) \\
	c_{\text{MPPI}}\left(z^R_t, u^R_t, T\right) &= (z^R_t - g^R)^\top Q_{\text{final}} \, (z^R_t - g^R) 
\end{align}
where $Q = \text{diag}([3.0, 3.0, 10.0, 0.0])$, $Q_{\text{final}} = Q / 5$, and $R = [2.0, 1.0]$.
\end{mdframed}
The first control from the control sequence is then executed before replanning again.
Similar to Algorithm~\ref{alg:hardware_accelerated_human_prediction_code}, computing the trajectories can be computed in parallel on a GPU.

Lastly, we note that these sequentially planners can also be extended to account for worst-case tracking error bounds using methods such as~\cite{herbert2017fastrack,siriya2025towards} to provide better safe guarantees as shown in~\cite{fridovich2020confidence,bajcsy2019scalable}. 

%% file: 4_experiments.tex
\section{EXPERIMENTS}
\begin{figure*}
\centering
\includegraphics[width=0.8\textwidth]{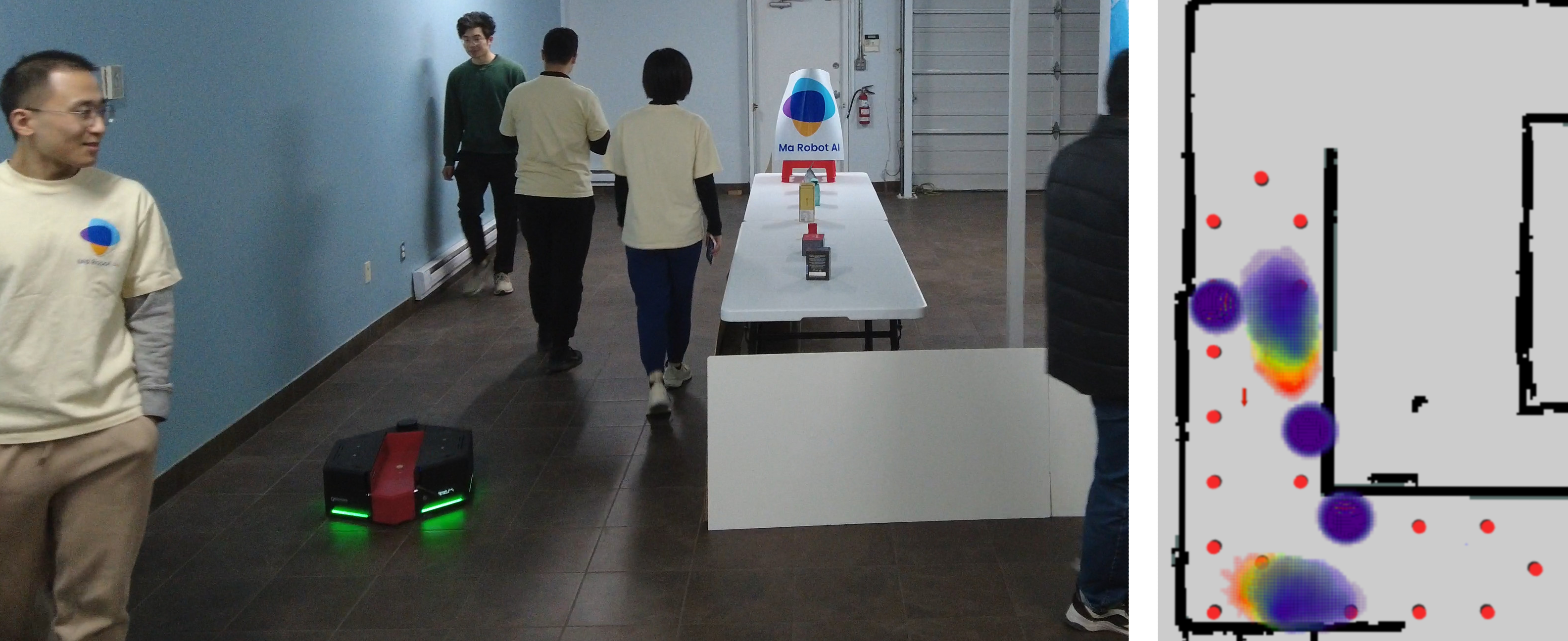} 
\caption{In the experiment, the humans are tasked to navigate in an $9.6 \times 5.4$ meter room as demonstrated in the left picture. The picture on the right shows an RVIZ display of the human predictions and their goals (red).}
\label{fig:human_walking}
\end{figure*}

We implement the proposed method in JAX~\cite{jax2018github} and conduct real-world experiments using the Robot Operating System 2 (ROS 2)~\cite{doi:10.1126/scirobotics.abm6074}.  
Our fully particle-based approach allows easy parallelization on hardware accelerators like GPUs using frameworks like JAX~\cite{jax2018github} to efficiently draw many particles to update of all probability distributions in real time.
The just-in-time (JIT) compilation in JAX further accelerates the algorithm's execution.
In Sec.~\ref{sec:numerical_simulation}, we measure the wall-clock time for running the algorithm and in Sec.~\ref{sec:real_world}, we evaluate the method in a real-world, human-aware robot navigation task to highlight the advantages of fast predictions in agile robot navigation among people. 

\subsection{Numerical Simulation}\label{sec:numerical_simulation}
We first begin by evaluating the wall-clock performance of our implementation across different configurations, comparing execution on the GPU versus the CPU, and assessing the impact of Just-In-Time (JIT) compilation.  
The experiments were conducted done on a desktop computer equipped with an AMD Ryzen 9 3950X processor, and a single NVIDIA GeForce RTX 2080 Ti GPU.
\begin{table}[h!]
\centering
\renewcommand{\arraystretch}{1.5} 
\begin{tabular}{|l|S[table-format=1.4]|S[table-format=1.4]|S[table-format=1.4]|S[table-format=1.4]|S[table-format=1.4]|}
\hline
\multicolumn{1}{|c|}{} & \multicolumn{5}{c|}{\textbf{Prediction Steps $T$}} \\
\cline{2-6}
\multicolumn{1}{|c|}{\textbf{Setting}} & \textbf{$2$} & \textbf{$4$} & \textbf{$6$} & \textbf{$8$} & \textbf{$10$} \\
\hline
CPU  & 0.130 & 0.642 & 1.242 & 1.865 & 2.493 \\
\hline
CPU w/ JIT & 0.050 & 0.097 & 0.139 & 0.189 & 0.229\\
\hline
GPU  & 2.826 & 2.943 & 3.022 & 3.192 & 3.253\\
\hline
GPU w/ JIT & 0.007 & 0.007 & 0.007 & 0.007 &  0.008\\
\hline\hline
\multicolumn{1}{|l|}{\textbf{Speed Up}} & \textbf{x 18} & \textbf{x 95}& \textbf{x 175}
& \textbf{x 257}
& \textbf{x 300}
\\
\hline
\end{tabular}
\caption{Wall clock (in seconds) for different device settings across varying number of prediction steps \( T \), and speed-ups of our method compared to CPU (without JIT), which represents a naive Python implementation of the particle-based approach. Note that such an approach would already be must faster than full Bayesian inference without the use of particles.}
\label{tab:wall_clock_times}
\end{table}

To measure the wall-clock of Alg.~\eqref{alg:hardware_accelerated_human_prediction_code}, we follow the same setting as in the running examples, a 2D kinematic as defined in Eq.~\eqref{eq:human_dynamics} with $n=8192$ samples, with the eventual predictions emplaced onto a $50 \times 50$ grid. 
Table~\ref{tab:wall_clock_times} summarizes the performance across multiple hardware and compilation settings with $\abs{\cb} = 5$, $\abs{\cg} = 10$ and $\abs{\cu} = 96$.
The wall-clock times are averaged over 5 runs.
When comparing implementations, the JIT methods are ``warm-started'' so that compilation time is not accounted for. 
This reflects the actual behavior of the algorithm when used for real world experiments, since the compilation only happens once during the start and can be cached between several executions.
On the backend, JAX's JIT compilations features are via~\cite{xla} that transforms written Python to optimized machine instructions (using C++).

As shown in the ``GPU w/ JIT" row of Table~\ref{tab:wall_clock_times}, the combination of executing the code on an GPU with JIT compilation enables human-prediction for a single human at $\sim$125 Hz.
This high-frequency prediction allows for a significantly larger time horizon or finer prediction time steps, which are critical for robust and adaptive planning.
In comparison to a naive Python implementation using running on the CPU (``CPU'' row of Table~\ref{tab:wall_clock_times}), we can achieve a 300 times speed up.

In contrast, using GPU parallelization without JIT (shown in the ``GPU" row of Table~\ref{tab:wall_clock_times}), results in the slowest prediction. 
We suspect this is due to operations only being executing on an GPU without any optimization or parallelization provided from the JIT compiler.

When executed on the CPU, the JIT provides a smaller improvement (shown in the ``CPU w/ JIT" row of Table~\ref{tab:wall_clock_times}). 
However, it lacks the parallel processing strength when compared to an GPU. 
Here, predictions of more than 4 time steps take at least 0.1s to compute, which can be too slow for robotic applications especially when the robot is also moving at considerable speeds. 
For example, if a human walks at a typical speed\footnote{\url{https://en.wikipedia.org/wiki/Preferred_walking_speed}} of $1.42$ m/s, with a robot moving at $1.0$ m/s, their relative distance may change by 0.24 m, which is a significant fraction of the average human body width.

Faster prediction directly improves planning quality in several ways.
First, it enables the robot to account for longer-term human trajectories, reducing the likelihood of unexpected collisions or suboptimal paths. 
Second, finer time steps provide more granular predictions, allowing the planner to react to subtle changes in human behavior.
Additionally, predictions for multiple-humans can be performed sequentially and efficiently, which will be shown in the following section.

\subsection{Safe Human-Aware Planning in the Real World}\label{sec:real_world}
To demonstrate the scalability of our approach, we conducted an experiment where a robot is tasked to patrol an occluded area while safely navigating around multiple humans with changing goals (see Fig.~\ref{fig:human_walking}).
A Quanser QBot was used to reach a set of fixed goals in sequence in an $9.6$m $\times$ $5.4$m room and is modeled using Eq.~\eqref{eq:robot_model}.
The robot's states were estimated using an Adaptive Monte-Carlo Localization method provided by the Nav2 package~\cite{macenski2020marathon}. 
The robot control's were bounded to allow a maximum speed of $1.1$ m/s and $1.0$ radians/s.

The human's $(x, y)$ positions were estimated using 3D object detection features obtain from an ZED 2 camera SDK.
The ZED2 camera was mounted in a fixed position in the room and tracked humans at frequency of 30 Hz. 
The humans were tasked to walk to random specified goal locations, distributed through the room.
During the experiments, our prediction method and robot do not have access to the true human navigational intents.

For both experiments, we used Alg.~\ref{alg:hardware_accelerated_human_prediction_code} (on the GPU w/ JIT) with $n=8192$ samples, $\abs{\cb} = 5$, $\abs{\cu} = 96$, $\Delta t = 0.5$, $T=6$ and varied the number of goals $\abs{\cg}$. 
The computation were done on a desktop computer equipped with an AMD Ryzen 9 5900X processor and a single NVIDIA GeForce RTX 4070 GPU.
For these environments, we used a grid size of $418$ $\times$ $195$ for computing trajectories using Anytime-A$^*$ and when emplacing the predictions onto a grid. 
This computation lead to resolution of $5$cm $\times$ $5$cm per grid cell.
If $k$ people were present in the environment, we would then track the states of each human to gather $\{z_{1:t}\}_{i=1}^k$ then compute their predictions sequentially. 
Finally, the predictions would be emplaced onto the same grid by taking a point-wise maximum to represent the union of probabilities.

During experiments, we found that the robot would act too conservatively when navigation around multiple people.
This was caused by the predictions occupying the majority of the free unoccupied space.
As a solution, we decided to dynamically change a human control's space if the camera detected them as stationary. 
Specifically, for a high velocity control $u'$, a large negative number used as the \emph{unknown state-control}, i.e, 
$Q_H(u', \mid z) = - \infty$.
This trick resulted in $\pi(u', z; \beta, g) \approx 0$ and is a commonly used in reinforcement learning~\cite{huang2020closer}.
Additionally, after a human leaves a goal, we reset their joint belief-goal distribution to be uniform.

As discussed in Sec.~\ref{sec:safe_traj_planners}, we integrated the predictions into both a global planner, ANA$^*$ and a local planner, MPPI.
Videos of the following experiments can be found at \url{https://www.youtube.com/playlist?list=PLUBop1d3Zm2sS2iI5TWN2wxQLjI_olpXd}.

\subsubsection{ANA$^*$}
For this experiment, we used ANA$^*$ (refer to Sec. \ref{sec:ana}) to generate a global path wtih $\abs{\cg} = 21$.
For path planning, we set a time-limit of $0.1$ seconds, and replanning was also done at 10 Hz.
If a collision-free path could not be found, the previous path is used.
We were able to compute these predictions at 11 Hz and the visualization of these predictions for 5 people in Fig.\ref{fig:human_walking}.
Compared to Table~\ref{tab:wall_clock_times}, a slower prediction frequency was achieved due to the need to predict the future occupancies of 5 people, compared to a single person, using a higher number of goal locations, and the additional overhead of visualizing the predictions.
However, the resulting prediction frequency were still faster than the planning frequency.
On average, the robot navigated around the humans at 0.7 m/s.
Since the predictions were integrated into the global planner, long-term collision paths could be found. 

\subsubsection{MPPI}
\begin{figure}[h]
    \centering
    \begin{subfigure}[b]{0.39\textwidth}
        \centering
        \includegraphics[width=\textwidth]{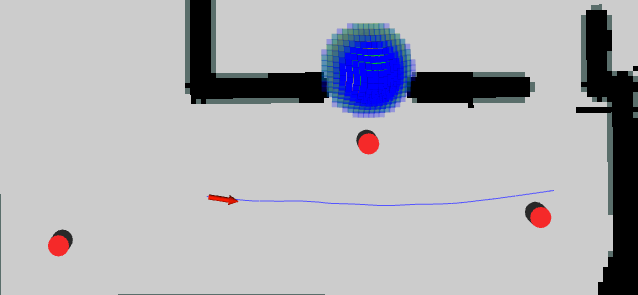} 
        \caption{$t=1$}
        \label{fig:t1}
    \end{subfigure}
        
    \begin{subfigure}[b]{0.39\textwidth}
        \centering
        \includegraphics[width=\textwidth]{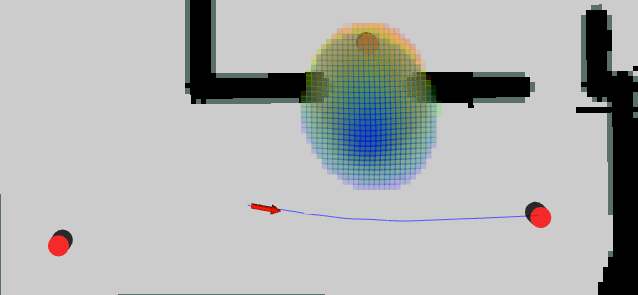} 
        \caption{$t=1.2$}
        \label{fig:t2}
    \end{subfigure}
    
    \begin{subfigure}[b]{0.39\textwidth}
        \centering
        \includegraphics[width=\textwidth]{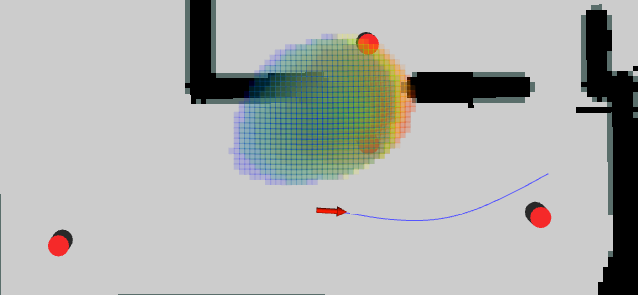} 
        \caption{$t=1.4$}
        \label{fig:t3}
    \end{subfigure}
    \caption{Evolution of Human Predictions and visualized MPPI trajectories over three snapshots of the experiment. The coloured visualizations indicate the future predicted occupancies. (a) A human reaches a goal. (b) As the human starts to walk towards another goal, the predictions change to be the mean of the two goals (red). As a result, the robot slightly quickly adjusts it's trajectory. (c) The human decides to walk to the lower left goal. Hence, the robot's plan largely remains the same.}
    \label{fig:mppi_stacked}
\end{figure}

In this section, we used the local planner MPPI (refer to~Sec.\ref{sec:mppi}) for robot navigation with $\abs{\cg} = 8$. 
In this setting, we used a planning horizon $T=40$ with a $\Delta_{\text{plan}} t =0.1$ and replan at 30 Hz.
We visualize the planned path and future human predictions every $0.2$s of part of the hardware demonstration in Fig.~\ref{fig:mppi_stacked}. 
Using the MPPI controller, the robot navigated at speeds up to 1.1 m/s.
Due to faster replanning, the robot could react to the change in human predictions much faster when comparison to ANA$^*$. 
When using Alg.~\ref{alg:hardware_accelerated_human_prediction_code} to predict the future states of two people along side the MPPI controller, the method produced predictions at 45 Hz. 
The frequency of prediction was slower since the MPPI controller was also running on the GPU, .

%% file: 5_conclusion.tex
\section{LIMITATIONS \& CONCLUSION}
We present a framework for confidence-aware human navigational prediction that operates in real-time for multiple humans. 
By parallelizing the Bayesian framework on a GPU, we achieve up to a 300 times speedup, enabling real-time prediction of the future human occupancies for multiple individuals. 
This approach is both efficient and scalable, as demonstrated in real-world experiments using local and global planners with up to five humans.

During experiments, we observed instances of ``near-misses'' between the robot and humans. 
We attribute these occurrences to tracking inaccuracies and model mismatches, which could be mitigated by integrating reachability-based methods such as FastTrack~\cite{herbert2017fastrack} to enhance safety guarantees. 
Additionally, the choice of the \emph{unknown state-control} function occasionally resulted in unrealistic predictions, such as predictions passing through obstacles. 
We also noted that humans often adjusted their behavior in response to the robot's movements, introducing further complexity to the prediction task.

For future work, we aim to address these challenges by incorporating reachability-based techniques, such as time-varying reach-avoid games~\cite{fisac2015reach}, into both the unknown state-control function and the planner. 
Furthermore, we plan to extend Bayesian inference to model group dynamics, which would reduce computational overhead when predicting the behavior of human groups. 
This extension would improve both efficiency and accuracy in multi-human environments.